\documentclass[twoside,twocolumn,9pt]{article}
\usepackage{extsizes}
\usepackage[super,sort&compress,comma]{natbib} 
\usepackage[version=3]{mhchem}
\usepackage[left=1.5cm, right=1.5cm, top=1.785cm, bottom=2.0cm]{geometry}
\usepackage{balance}
\usepackage{mathptmx}
\usepackage{sectsty}
\usepackage{graphicx} 
\usepackage{lastpage}
\usepackage[format=plain,justification=justified,singlelinecheck=false,font={stretch=1.125,small,sf},labelfont=bf,labelsep=space]{caption}
\usepackage{float}
\usepackage{fancyhdr}
\usepackage{fnpos}
\usepackage{ulem}
\usepackage[english]{babel}
\addto{\captionsenglish}{%
  
}
\usepackage{array}
\usepackage{amsmath}
\usepackage{droidsans}
\usepackage{charter}
\usepackage[T1]{fontenc}
\usepackage[usenames,dvipsnames]{xcolor}
\usepackage{setspace}
\usepackage[compact]{titlesec}
\usepackage{hyperref}

\usepackage{epstopdf}

\definecolor{cream}{RGB}{222,217,201}

\begin{document}

\pagestyle{fancy}
\thispagestyle{plain}
\fancypagestyle{plain}{
\renewcommand{\headrulewidth}{0pt}
}

\makeFNbottom
\makeatletter
\renewcommand\LARGE{\@setfontsize\LARGE{15pt}{17}}
\renewcommand\Large{\@setfontsize\Large{12pt}{14}}
\renewcommand\large{\@setfontsize\large{10pt}{12}}
\renewcommand\footnotesize{\@setfontsize\footnotesize{7pt}{10}}
\makeatother

\renewcommand{\thefootnote}{\fnsymbol{footnote}}
\renewcommand\footnoterule{\vspace*{1pt}%
\color{cream}\hrule width 3.5in height 0.4pt \color{black}\vspace*{5pt}} 
\setcounter{secnumdepth}{5}

\makeatletter 
\renewcommand\@biblabel[1]{#1}            
\renewcommand\@makefntext[1]%
{\noindent\makebox[0pt][r]{\@thefnmark\,}#1}
\makeatother 
\renewcommand{\figurename}{\small{Fig.}~}
\sectionfont{\sffamily\Large}
\subsectionfont{\normalsize}
\subsubsectionfont{\bf}
\setstretch{1.125} 
\setlength{\skip\footins}{0.8cm}
\setlength{\footnotesep}{0.25cm}
\setlength{\jot}{10pt}
\titlespacing*{\section}{0pt}{4pt}{4pt}
\titlespacing*{\subsection}{0pt}{15pt}{1pt}

\fancyfoot{}
\fancyfoot[LO,RE]{\vspace{-7.1pt}\includegraphics[height=9pt]{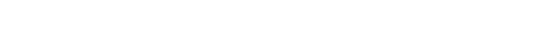}}
\fancyfoot[CO]{\vspace{-7.1pt}\hspace{13.2cm}\includegraphics{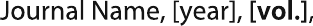}}
\fancyfoot[CE]{\vspace{-7.2pt}\hspace{-14.2cm}\includegraphics{head_foot/RF}}
\fancyfoot[RO]{\footnotesize{\sffamily{1--\pageref{LastPage} ~\textbar  \hspace{2pt}\thepage}}}
\fancyfoot[LE]{\footnotesize{\sffamily{\thepage~\textbar\hspace{3.45cm} 1--\pageref{LastPage}}}}
\fancyhead{}
\renewcommand{\headrulewidth}{0pt} 
\renewcommand{\footrulewidth}{0pt}
\setlength{\arrayrulewidth}{1pt}
\setlength{\columnsep}{6.5mm}
\setlength\bibsep{1pt}

\makeatletter 
\newlength{\figrulesep} 
\setlength{\figrulesep}{0.5\textfloatsep} 

\newcommand{\topfigrule}{\vspace*{-1pt}%
\noindent{\color{cream}\rule[-\figrulesep]{\columnwidth}{1.5pt}} }

\newcommand{\botfigrule}{\vspace*{-2pt}%
\noindent{\color{cream}\rule[\figrulesep]{\columnwidth}{1.5pt}} }

\newcommand{\dblfigrule}{\vspace*{-1pt}%
\noindent{\color{cream}\rule[-\figrulesep]{\textwidth}{1.5pt}} }

\makeatother

\twocolumn[
  \begin{@twocolumnfalse}
{\includegraphics[height=30pt]{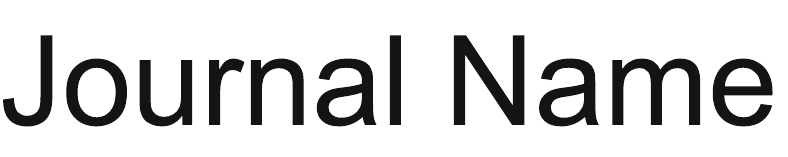}\hfill\raisebox{0pt}[0pt][0pt]{\includegraphics[height=55pt]{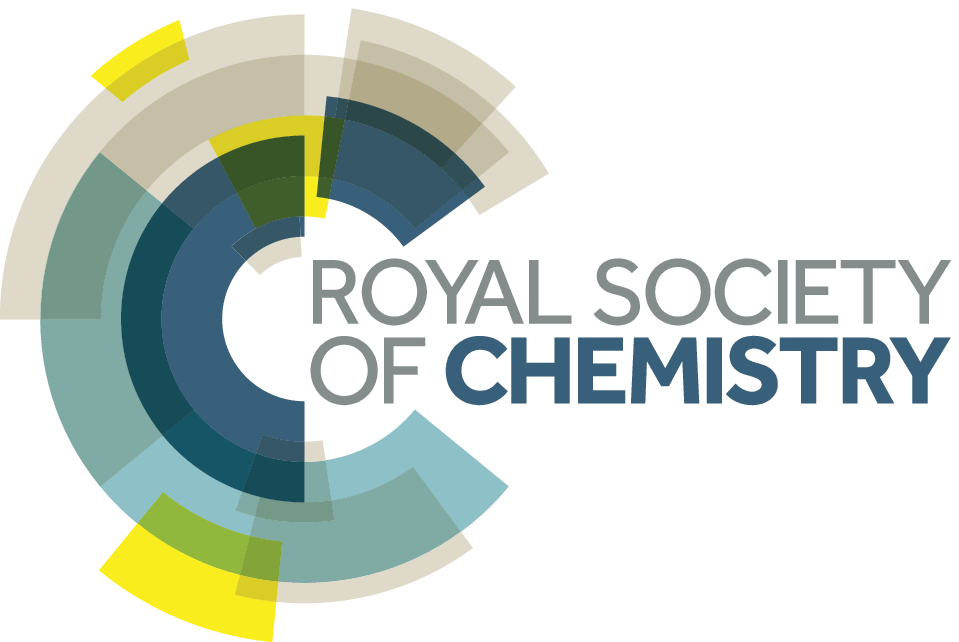}}\\[1ex]
\includegraphics[width=18.5cm]{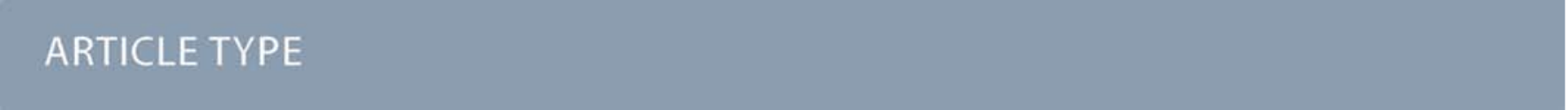}}\par
\vspace{1em}
\sffamily
\begin{tabular}{m{4.5cm} p{13.5cm} }

\includegraphics{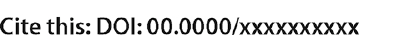} & \noindent\LARGE{\textbf{Fragment-based molecular generative model with high generalization ability and synthetic accessibility.}} \\
\vspace{0.3cm} & \vspace{0.3cm} \\

 & \noindent\large{Seonghwan Seo,\textit{$^{{\dag}a}$} Jaechang Lim,$^{\ast}$$^{\dag}$\textit{$^{a}$} and Woo Youn Kim$^{\ast}$\textit{$^{a,b,c}$}} \\

\includegraphics{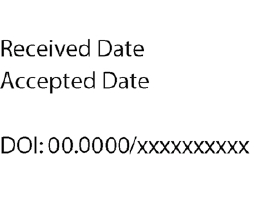} & \noindent\normalsize{Deep generative models are attracting great attention for molecular design with desired properties. Most existing models generate molecules by sequentially adding atoms. This often renders generated molecules with less correlation with target properties and low synthetic accessibility. Molecular fragments such as functional groups are more closely related to molecular properties and synthetic accessibility than atoms. Here, we propose a fragment-based molecular generative model which designs new molecules with target properties by sequentially adding molecular fragments to any given starting molecule. A key feature of our model is a high generalization ability in terms of property control and fragment types. The former becomes possible by learning the contribution of individual fragments to the target properties in an auto-regressive manner. For the latter, we used a deep neural network that predicts the bonding probability of two molecules from the embedding vectors of the two molecules as input. The high synthetic accessibility of the generated molecules is implicitly considered while preparing the fragment library with the BRICS decomposition method. We show that the model can generate molecules with the simultaneous control of multiple target properties at a high success rate. It also works equally well with unseen fragments even in the property range where the training data is rare, verifying the high generalization ability. As a practical application, we demonstrated that the model can generate potential inhibitors with high binding affinities against the 3CL protease of SARS-COV-2 in terms of docking score.} \\

\end{tabular}

 \end{@twocolumnfalse} \vspace{0.6cm}
  ]

\renewcommand*\rmdefault{bch}\normalfont\upshape
\rmfamily
\section*{}
\vspace{-1cm}

\footnotetext{\textit{$^{a}$~HITS incorporation, 124, Teheran-ro, Gangnam-gu, Seoul, Republic of Korea; Email: jaechang@hits.ai}}
\footnotetext{\textit{$^{b}$~Department of Chemistry, KAIST, 291 Daehak-ro, Yuseong-gu, Daejeon 34141, Republic of Korea; Email: wooyoun@kaist.ac.kr}}
\footnotetext{\textit{\textsuperscript{c}KI for Artificial Intelligence, KAIST, 291 Daehak-ro, Yuseong-gu, Daejeon 34141, Republic of Korea}}

\footnotetext{\dag~These authors contributed equally to this work.}



\section{Introduction}

The design of molecules with desired properties is at the heart of chemistry. It is challenging to design new molecular structures with the simultaneous control of multiple properties due to complicated structure-property relationships. Deep learning-based molecular design has attracted great attention as a new strategy for various applications including drug design.\cite{Sanchez-Lengeling2018, Elton2019, Chen2021, Schneider2020} The so-called deep generative models aim to precisely control multiple properties while navigating the vast chemical space. That becomes possible by learning the structure-property relationships directly from raw data implying both structure and property information on diverse molecules. Recent studies demonstrated that the generative models can be applied to designing drug candidates in early-stage drug discovery such as hit generation and lead optimization. For example, Zhavoronkov et al. designed focused molecules as inhibitors against discoidin domain receptor family member 1, and their potency and physicochemical properties measured by experiments indeed satisfied conditions as lead candidates. \cite{Zhavoronkov2019} 

The architectures of various deep generative models reply on molecular representations employed as input for molecular structures. Language models with SMILES molecular representation have been widely used.\cite{Segler2018focus} 
The language models are trained to construct new SMILES strings by sequentially adding new characters to a given piece of SMILES string. Then, the models can generate novel molecules not in the training set by exploring the chemical space learned from the training. Variational autoencoder (VAE) is also a popular architecture.\cite{Gomez-Bombarelli2018} In VAE, the encoder converts a given molecular representation as input to an embedding vector in the latent space, and the decoder recovers the original molecule from the latent vector. After training, the VAE model can generate new molecules by decoding latent vectors sampled from the resulting latent space. The controlled sampling of latent vectors allows us to manipulate the structural diversity and similarity of the generated molecules. It is also possible to control the molecular properties of the generated molecules by adopting additional training strategies such as reinforcement learning\cite{Olivecrona2017,Popova2018,Guimaraes2017}, transfer learning\cite{Segler2018focus}, and conditional generation\cite{Lim2018,Kang2019}.

Graph-based deep generative models can improve the quality of molecule generation.\cite{Jin2018,Li2018,Assouel2018,Liu2018,You2018,li2018multi} The molecular graph can naturally represent chemical validity and molecular similarity, resulting in superior feature extraction compared to the SMILES representations. 
Jin et al.\cite{Jin2018} and Li et al.\cite{li2018multi} reported that graph-based deep generative models show better validity, uniqueness, and novelty of generated molecules than those of the SMILES-based models, indicating that the learned probability distribution is closer to the true distribution of existing molecules. 

Moreover, the graph-based models can be readily modified for specialized purposes through specifically controlling molecular graph structures because their nodes and edges directly correspond to atoms and chemical bonds. Lim et al.\cite{Lim2020} and Li et al.\cite{Li2020scaffold} proposed scaffold-based molecule generation algorithms for early stage drug discovery such as hit-to-lead and lead optimization in which molecular structures can be adjusted without changing a designated core scaffold. 
Imrie et al. proposed a linker generation model while conserving given fragments and their coordinates, which can assist to combine small molecules in fragment-based drug discovery.\cite{Imrie2020} Besides, several models have been proposed to improve the synthetic feasibility of generated molecules, which is practically very important.\cite{Bradshaw2019,krishna2020} Bradshaw et al. proposed a reaction-based generative model, which sequentially chooses reactants and appropriate reaction templates from predefined candidates and reaction templates.\cite{Bradshaw2019}

Despite the promising results of previous models, their common design strategy that sequentially adds atoms and bonds would be chemically less intuitive. Human experts mostly perceive a molecule as a connected set of functional substructures rather than a simple assembly of atoms. In terms of designing molecules, this conceptual perception is more practical because molecular properties are finely tuned by tailoring specific functional groups. This fragment-based molecular design is also advantageous as considering the synthetic feasibility of generated molecules. This can be done, for example, by preparing synthetically accessible fragments using known reaction templates and letting the model learn the implied synthetic validity from the resulting data. In this regard, we propose a fragment-based molecular generative model that aims to design new molecules by sequentially adding molecular fragments to any given starting molecule. In a training phase, the model learns to recover original molecules by adding molecular fragments to an arbitrarily given core structure. In a generation phase, it predicts a possible molecular fragment and corresponding atom pairs for making a bond between the fragment and the core structure. At the end, a novel molecule with desired properties can be obtained by repeating the process. 

We expect that the sequential addition of molecular fragments helps the model learn how each substructure affects the molecular properties, rather than simply memorizing the relationship between whole molecular structures and their properties in an end-to-end fashion. That is, the model can learn how to select appropriate fragments and bind them with given core molecules to achieve target properties.
Moreover, learning the contribution of each fragment to target properties can encourage the model to produce novel molecules even with rare property values in the training set. 
In this perspective, the sequential addition of molecular fragments is more beneficial than the sequential addition of atoms for learning the structure-property relationship with high generalization ability, because molecular properties are more correlated with functional molecular fragments than individual atoms.  

Efficient processing of numerous molecular fragments is a key challenge for achieving the high diversity of generated molecules in fragment-based deep generative models. 
For instance, one can obtain more than 70,000 unique molecular fragments from 500,000 molecules randomly selected from any molecular database with the BRICS decomposition method\cite{Degen2008}. 
Formulating the fragment selection problem from a library into a classification task may provoke serious limitations. First, representing the whole set of fragments as a single vector and training the classification model are computationally inefficient unless using a small number of fragments. However, the use of a small number of fragments reduces the diversity of generated molecules. Second, the model must be retrained whenever a new fragment is added to the library. 
We solve these problems by splitting the fragment selection process into two steps.  
We first sample a fragment randomly from a predefined library. Then, we determine whether the sampled fragment will be added to a given core molecule, which can be done using a deep neural network that predicts the probability of connecting between two molecules. The neural network can take any two molecules as an embedding vector obtained by encoding the molecules using another deep neural network, so one can add new fragments in the library without retraining the model. This strategy allows us to handle an unlimited number of fragments in theory while maintaining high computational efficiency.
To our best knowledge, there are a few fragment-based molecular generative models. 
Podda et al. proposed a language model which sequentially generates fragments and combining them into a single molecule.\cite{podda2020} They could achieve the high validity and uniqueness of generated molecules. Yang et al. developed a reinforcement learning model that sequentially adds fragments to a given core molecule to improve the binding affinity of the resulting molecule to a target protein.\cite{Yang2021} The example study in the work showed a possibility of designing potential drug candidates with strong binding to the target.
Despite the conceptual advance of these models and the encouraging results, the two models have fundamental limitations in dealing with diverse fragments. Yang et al. sampled fragments from a predefined library which contains only 66 fragments. Podda et al. explicitly considered only a small number of frequent molecular fragments in a dataset. Furthermore, these models cannot accept novel fragments that are not in the training set, because they used fixed libraries. Chen et. al. solved the limitations by representing fragments with latent vectors and searching fragments in the resulting latent space.\cite{Chen2021} However, sampling fragments from the latent space does not guarantee the synthetic accessibility of generated molecules especially when the fragments are not readily available.
In contrast, our model has no such limitations as explained above.

\section{Method}
\begin{figure*}
 \centering
 \includegraphics[height=17cm]{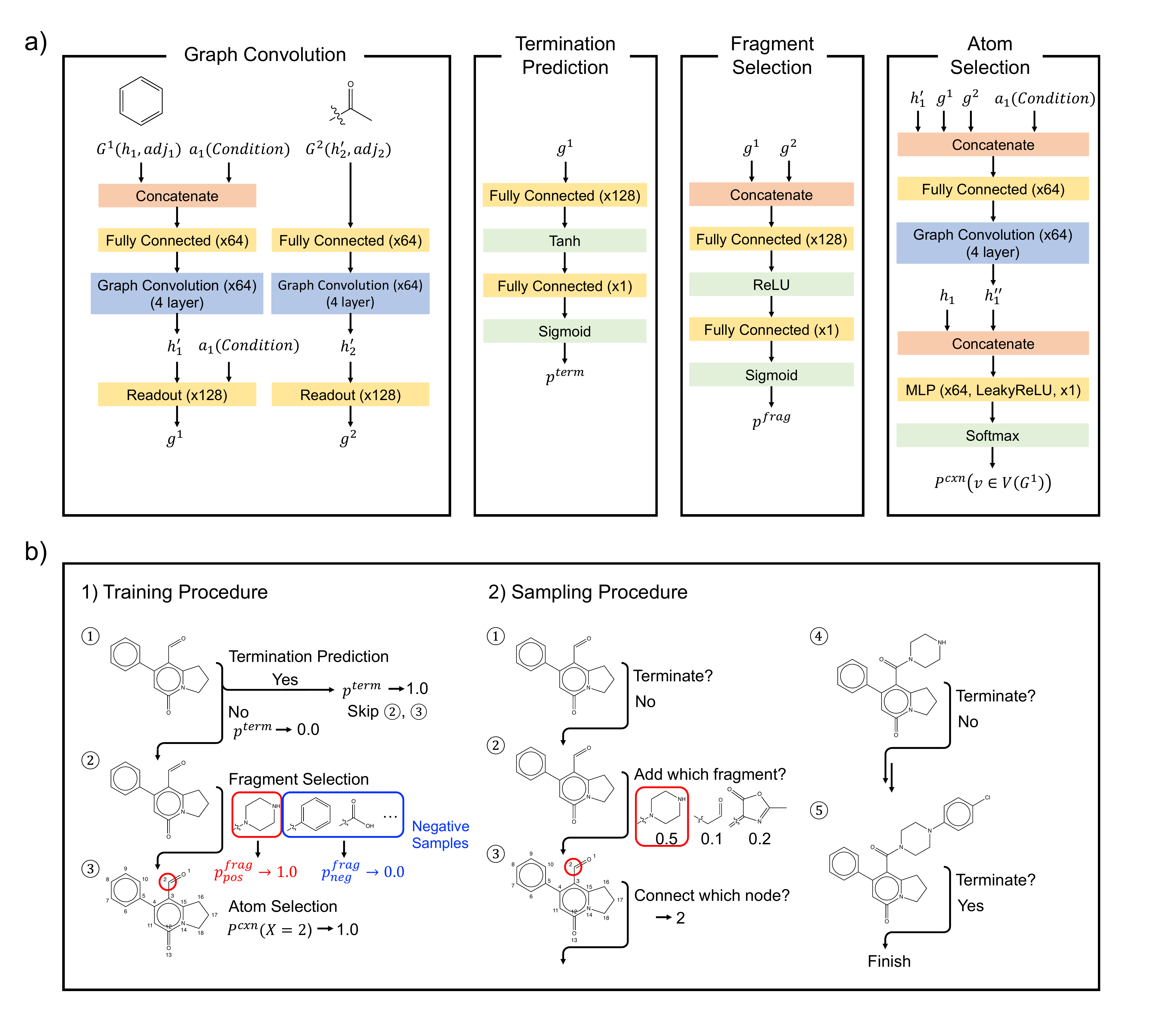}
 \caption{a) The schematic representation of the model. b-1) the training procedure and b-2) the sampling procedure of the model.}
 \label{fig:scheme}
\end{figure*}

Our goal is to generate functional molecules for a specific purpose by sequentially adding molecular fragments to any  given core molecule as input until satisfying desired properties. To this end, our model has three sub-modules: a fragment selection module, an atom selection module, and a termination module. The fragment selection module predicts an appropriate molecular fragment to be added. The atom selection module finds an atom pair for making a bond between the predicted fragment and the core molecule: one from the predicted fragment and the other from the core molecule. The termination module determines whether the generation process should be terminated or repeated. Fig.~\ref{fig:scheme} schematically shows the model architecture and the process of the training and generation. We describe the details of each sub-module and the processes in the following subsections.

\subsection{Dataset}
An important prerequisite for developing fragment-based deep generative models is the preparation of an appropriate fragment library. The library should contain all fragments of the molecules in the training set, preferably many molecules, so that the model can learn chemical diversity. There would be various definitions of molecular fragments such as BRICS\cite{Degen2008}, Recap\cite{Lewell1998}, Bemis-Murcko\cite{Bemis1996}, and etc. In principle, our proposed model works with any definition. In this work, we adopted the BRICS decomposition. The algorithm of the BRICS decomposition is breaking the covalent bonds which correspond to predefined SMARTS strings. The resulting fragments have labels on their atoms indicating whether the formation of chemical bonds at the atom is possible. These labels help the model generate synthetically more feasible molecules in the generation process. To construct a dataset for training and test, we applied the BRICS decomposition to 583,876 molecules chosen randomly from the MolPort library, resulting in 70,100 unique fragments.


\subsection{Fragment selection}
To handle many fragments efficiently in the dataset, we formulated the fragment selection task into a matching problem of two molecular graphs for connection. The selection module takes two molecular graphs $G^1$=($V^1$,$E^1$) and $G^2$=($V^2$,$E^2$) as input, where $V$ and $E$ denote a set of nodes (or atoms) and edges (or bonds) in a given graph (molecule), and then predicts the probability of binding $G^1$ and $G^2$. Each graph $G$ has nodes $v_i \in V$ and edges $e_{ij} \in E$, where $i$ and $j$ denote node indices. In our work, $G^1$ and $G^2$ are the molecular graph of a core molecule given as input or from the previous step and a fragment sampled randomly from the fragment library, respectively. The model is based on a graph convolution network defined as follows:
\begin{equation} \label{eq:eq1}
    h_i' = \phi_1(h_i\|a_i)
\end{equation}
\begin{equation} \label{eq:eq2}
    h_i'' = ReLU(\sum_{j \in N_i} \phi_2(h_j'))
\end{equation}
\begin{equation} \label{eq:eq3}
    c_i = \sigma(\phi_3(h_i''|h_i'))
\end{equation}
\begin{equation} \label{eq:eq4}
    h_i''' = c_{i}h_{i}'+(1-c_{i})h_{i}'',
\end{equation}
where $h_i$ is a n-dimensional embedding vector of $v_i$, $ReLU$ is a ReLU activation funtion, $\sigma$ is a sigmoid activation function, $N_i$ is the neighboring nodes of the $i$-th node, $a_i$ is an additional attribute vector, and $\phi_1$, $\phi_2$, and $\phi_3$ are fully connected layers, respectively. The output embedding vector $h_i'''$ of a graph convolution layer is used as an input embedding vector of the next graph convolution layer ($h_i:=h_i'''$). The purpose of eqn \ref{eq:eq1} is to embed an attribute vector $a_i$ with $h_i$ into a vector and to achieve enough expressive power by adopting a linear layer. Any attribute such as atomic or molecular properties can be included in $a_i$. Then, the model learns the joint probability distribution of the given molecule and its properties. After applying the graph convolution layers several times, we obtain a graph vector $g$ of $G$ from the weighted summation of the final embedding vectors; 
\begin{equation} \label{eq:eq5}
    g_{sum} = \sum_i \sigma(\phi_4(h_i))\phi_5(h_i)
\end{equation}
\begin{equation} \label{eq:eq6}
    g = \phi_6(g_{sum}|a_g),
\end{equation}
where $\phi_4$, $\phi_5$, and $\phi_6$ are fully connected layers, and $a_g$ is an additional attribute vector. By applying the graph convolution layers to each of $G^1$ and $G^2$, we obtain the two respective graph vectors $g^1$ and $g^2$. To train the conditional model, we incorporated molecular properties into the attribute vectors of $G^1$. The probability value for fragment selection $p^{frag}$ is evaluated as a function of $g^1$ and $g^2$ as follows.
\begin{equation}
    p^{frag} = \sigma(\phi_7((g^1\|g^2))),
\end{equation}
where $\phi_7$ is a neural network made of fully connected layers.

The training dataset made by the BRICS decomposition intrinsically includes only positive samples, where "positive" means that fragments of a molecule can be properly added to the other fragments of the same molecule prepared by the decomposition method. To make the model select such fragments over others, we also need to train the model with namely the negative samples which are unlikely to be added. To prepare the negative samples, we randomly chose BRICS fragments for each positive sample as negatives. 
This training strategy, called the negative sampling, is often used in the Word2Vec model in the natural language processing.\cite{Mikolov2013}
The model is trained to predict the fragment probability $p^{frag}$  as 1.0 for the positive samples and 0.0 for the negative samples. We used the binary cross-entropy loss for this task. The objective function $J_{fragment}$ of the fragment selection module is as follows:
\begin{equation}
    J_{fragment} = \log p^{frag}_{pos} + \frac{1}{N}\sum_{i}^{N}\log\left(1-p_{neg, i}^{frag}\right),
\end{equation}
where $N$ is the number of negative samples per each positive sample, and $p^{frag}_{pos}$ and $p^{frag}_{neg}$ are predicted probability values for positive samples and negative samples, respectively.

\subsection{Atom selection}
The atom selection module predicts the bonding probability between $v_i^1 \in V(G^1)$ and $v_j^2 \in V(G^2)$. The atom for the bonding in the selected fragment $G^2$ is already labeled when it is prepared by the BRICS decomposition. Hence, we need to choose the counterpart atom from $G^1$. The atom selection module accepts output node embedding vectors from the fragment selection module and applies the graph convolution layers to them. In this process, we assign the graph vectors $g^1$ and $g^2$ as the attribute vectors $a_1$ in $G^1$, and $a_2$ in $G^2$, respectively, to help the model effectively learn features related to the two graphs. For training the conditional model, we also incorporate molecular properties as the attribute vectors in graph convolution layers (eqn (\ref{eq:eq1})--(\ref{eq:eq4})). After applying the graph convolution layers, the connection probability $p^{cxn}_i$ of $v_i^1 \in V(G^1)$ is calculated using fully connected layers with a softmax layer at the end. The model is trained to predict $p^{cxn}_i$ as 1 for a positive atom and 0 for negative atoms. We used the cross-entropy loss for this task. The objective function $J_{selection}$ is as follow:
\begin{equation}
    J_{selection} = \frac{1}{|V(G^1)|}\sum_{v_{i}^{1}\in{V(G^1)}}{}y^{cxn}_i\log{p^{cxn}_i},
\end{equation}
where $y^{cxn}_i$ is the true label of $v_i$ indicating whether the atom is positive or negative.

\subsection{Termination prediction}
The termination prediction module predicts the probability of terminating the generation process. The module produces a graph vector $g^1$ by applying the graph convolution layers ((\ref{eq:eq1})--(\ref{eq:eq4})) and weighted summation of embedding vectors ((\ref{eq:eq5})--(\ref{eq:eq6})). Then, the termination probability can be evaluated by applying fully connected layers $\phi_8$ with the sigmoid activation function to the graph vector $g^1$, which is given by 
\begin{equation}\label{eq:eq10}
    p^{term} = \sigma(\phi_8(g^1)).
\end{equation}
In the training process, the module is trained to predict the termination probability $p^{term}$ as 1.0 when the original molecules are recovered and otherwise as 0.0.
The objective function $J_{termination}$ is given by
\begin{equation}
    J_{termination} = y^{term}\log{p^{term}} + (1-y^{term})\log{(1-p^{term})},
\end{equation}
where $y^{term}$ is the termination label.

\subsection{Molecule generation after training the model}
The model accepts a starting molecule as input to generate a larger molecule by adding fragments to it. If the starting molecule is not given, a fragment is randomly selected from the fragment library as a starting molecule. The conditional molecule generation additionally needs target properties as input. 
The generation process begins to predict the termination probability of the starting molecule. Then, the termination sign is sampled in proportion to the termination probability given by eqn \eqref{eq:eq10}. If not terminating, the model executes the fragment selection and the atom selection modules subsequently. In the fragment selection step, the model randomly samples a set of BRICS fragments proportional to their populations in the training. This stochastic sampling enhances the efficiency of the generation process. The number of BRICS fragments in each sampling is a hyper-parameter, and here we sampled 2,000 BRICS fragments at each time. After sampling the BRICS fragments, the fragment selection module predicts the matching probability of every fragment for addition and stochastically selects one of them in proportion to its predicted probability. Then, the atom selection module predicts the connection probability of all possible atoms in the starting molecule. We stochastically choose one atom in proportion to the predicted probability. Finally, we connect the labeled atom of the fragment with the chosen atom. These procedure repeats until the termination sign is on.

Fig.~\ref{fig:scheme}--a) shows the hyper-parameters of the model.
The source code and dataset in this work are available at \url{https://github.com/jaechang-hits/FMGM-pytorch}.


\section{Results and discussion}
\subsection{Controlling molecular properties of generated molecules}

\begin{figure}
 \includegraphics[height=12cm]{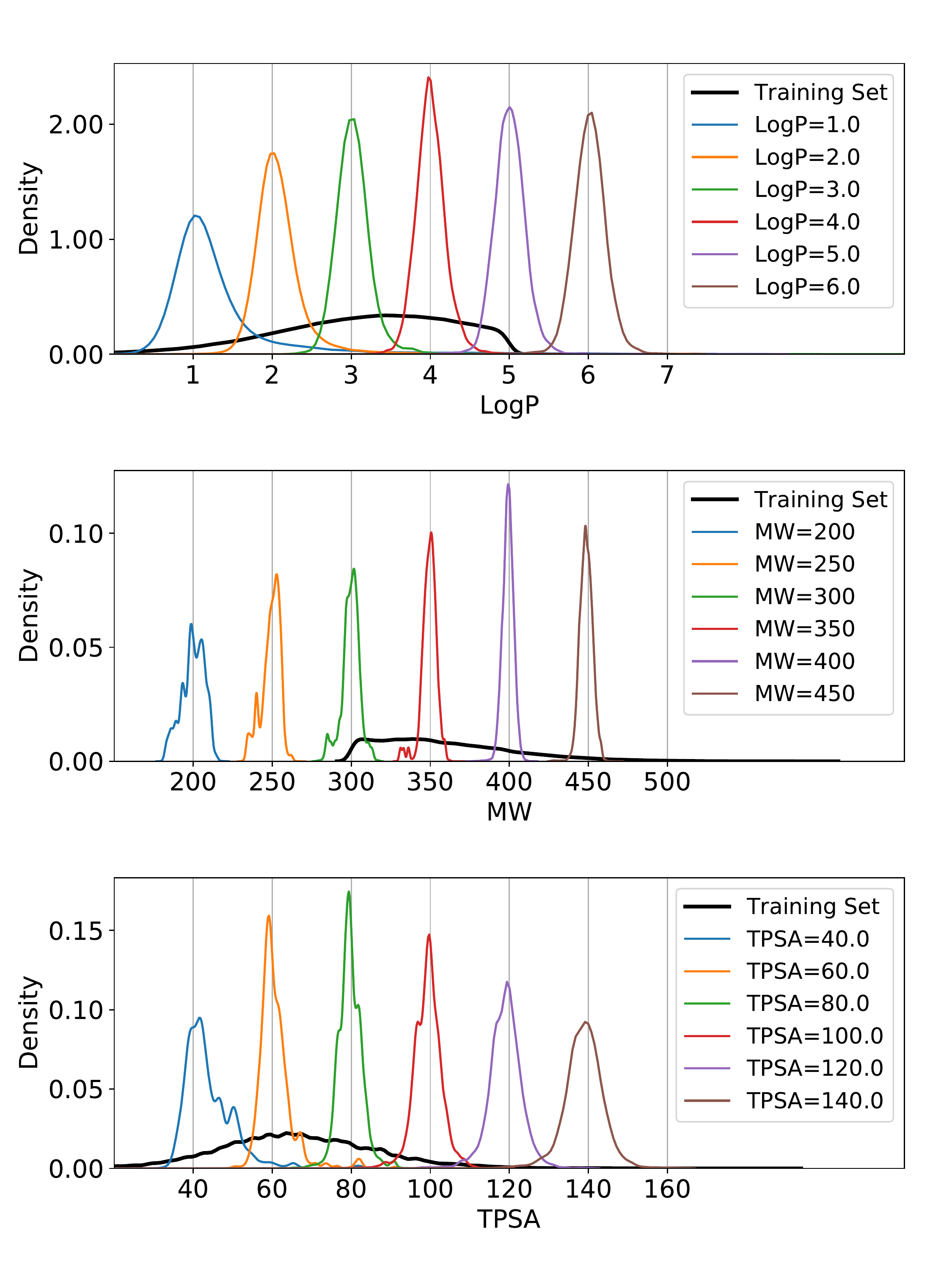}
 \caption{The property distributions of the molecules generated by the model conditioned on LogP, MW, and TPSA, respectively. The black line indicates the property distribution of the training set, while the color lines denote those of the generated molecules with various target properties.}
 \label{fig:single_property}
\end{figure}

One criterion of assessing the performance of deep generative models for molecule generation is comparing the designated target property and the actual property of generated molecules. For this purpose, we trained three instances of the model with three properties: molecular weight (MW), LogP, and TPSA. We used RDKit to calculate the three molecular properties. After training, we generated 100 molecules for each of the 100 random starting molecules from the test dataset.
For the property control task, we only tested the molecule generation starting from a given molecular fragment because it is more challenging than \textit{de novo} design due to the constraint imposed by the fixed starting molecules. 
We repeated the process for the three target properties. For MW, we used only small starting molecules whose MW is smaller than the target MW because it is unphysical to reduce MW by adding molecular fragments. 
Fig.~\ref{fig:single_property} shows the property distribution of the generated molecules with different target properties. Their peak positions are at the target property, indicating that the model successfully learned the structure-property relationship in terms of controlling the molecular properties. Even in regions where training data points are sparse, the distribution is as sharp as in data-rich regions. It was possible because the model learns the contribution of each fragment to the target molecular properties instead of learning the end-to-end mapping between the whole molecular structure and its property, as mentioned in the introduction.

Also, we calculated the validity and uniqueness of the generated molecules. The exact definitions of the three metrics were obtained from ref~\citenum{Lim2020}. In practice, we used RDKit\cite{landrum2006rdkit} to determine the validity of the generated molecules. These metrics indicate not only the efficiency of molecule generation but also the quality of learned distributions. Table \ref{tab:single_property} shows the two metric values of unconditional and conditional generation with respect to the kind of molecular properties and the property value. The "mean" indicates the absolute average difference between the target property and the actual property of the generated molecules. The "std" indicates the standard deviation of the absolute differences. The model shows the high values in terms of the two metrics regardless of the molecular properties and the target values except for a few extreme cases, for example, when MW is 200. 
Despite the restrictions imposed by the fixed starting molecules, the high uniqueness means that the model achieves good generalization ability rather than the memorization of the hidden patterns in the training set. 
Also, the model improved the validity and uniqueness from those of the previous deep graph generative models. We note that each model has been tested under different conditions. Therefore, it is meaningful to compare our model with the scaffold-based GGM which generates new molecules by sequentially adding atoms and bonds to a given scaffold. Indeed, our model outperformed the scaffold-based GGM model. In addition, the model shows the high validity and uniqueness for both generation scenarios starting with and without fixed starting molecules. 

\begin{table}[]
\caption{The validity, uniqueness, mean, and standard deviation (std) of the conditional and unconditional generation. The "mean" indicates the absolute average difference between the target property and the actual property of the generated molecules. ours (Unconditional)$^a$ and ours (Unconditional)$^b$ mean the molecule generation modes with and without fixed starting molecules, respectively}
\begin{tabular}{ccccc}
          & Validity & Uniqueness & Mean   & Std  \\ \hline
ours (LogP=1) & 97.9     & 92.1       & 1.36   & 0.92  \\
ours (LogP=2) & 98.1     & 89.8       & 2.12   & 0.52 \\
ours (LogP=3) & 98.3     & 94.1       & 3.04   & 0.32 \\
ours (LogP=4) & 98.5     & 95.0       & 4.01   & 0.21  \\
ours (LogP=5) & 98.4     & 98.2       & 5.01   & 0.21 \\
ours (LogP=6) & 98.0     & 99.5       & 6.02   & 0.24 \\ \hline
ours (MW=200)    & 93.2     & 29.0       & 199.9 & 7.1 \\
ours (MW=250)    & 95.4     & 44.0       & 249.1 & 5.8 \\
ours (MW=300)    & 97.5     & 53.8       & 299.6 & 5.6 \\
ours (MW=350)    & 97.6     & 76.5       & 349.3 & 4.7 \\
ours (MW=400)    & 97.2     & 93.2       & 399.2 & 3.5 \\
ours (MW=450)    & 97.0     & 96.2       & 448.6 & 4.0 \\ \hline
ours (TPSA=40)   & 71.3     & 91.4       & 43.94  & 6.20 \\
ours (TPSA=60)   & 94.3     & 91.5       & 60.81  & 4.35 \\
ours (TPSA=80)   & 97.2     & 94.9       & 79.60  & 3.07 \\
ours (TPSA=100)  & 97.3     & 97.1       & 99.31  & 3.64 \\
ours (TPSA=120)  & 96.6     & 99.2       & 118.91 & 4.53 \\
ours (TPSA=140)  & 96.1     & 99.6       & 138.67 & 6.10 \\ \hline
ours (Unconditional)$^a$ & 98.9 & 91.5 & - & - \\
ours (Unconditional)$^b$ & 97.5 & 100.0 & - & - \\
Scaffold-based GGM\cite{Lim2020} & 96.5 & 85.6 & - & - \\
DeepScaffold\cite{Li2020scaffold} & 82.5 & - & - & - \\
GraphVAE\cite{simonovsky2018graphvae} & 55.7 & 87.0 & - & - \\
MolGAN\cite{DeCao2018} & 98.1 & 10.4 & - & - \\ \hline
\end{tabular}
 \label{tab:single_property}

\end{table}

\begin{figure}
 \includegraphics[height=10cm]{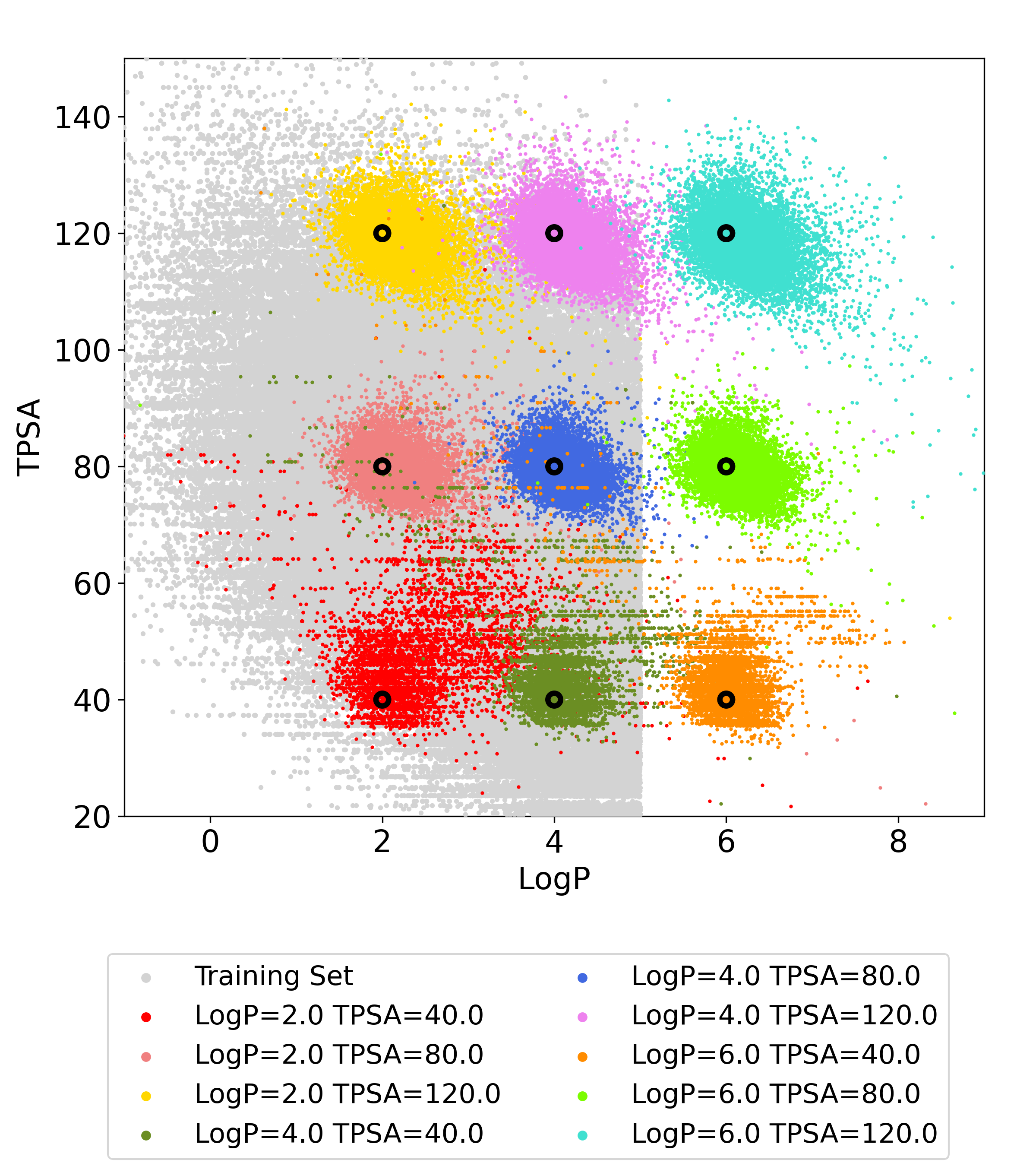} 
 \caption{The LogP and TPSA values of the molecules generated by the model conditioned on both LogP and TPSA. The gray dots are for the training set, while the color dots are for the generated molecules. We used RDKit to calculate the LogP and TPSA of the molecules.} 
 \label{fgr:noraml_multi}
\end{figure}

We further tested whether the model can control multiple molecular properties of generated molecules. 
For that purpose, we trained the instance of the model with LogP and TPSA, and generated 100 molecules for each of the same 100 starting molecules used in the previous experiment. Fig.~\ref{fgr:noraml_multi} shows the property distributions of the molecules as a function of the target values. The gray points indicate molecules in the training set. Though the distributions are more spread than the case of the single property control, their peak positions are at the target values. It supports that the model can correctly learn the structure-multiple properties relationship. In addition, the model performs well for the extreme targeted values around TPSA of 120 and LogP of 6, though the training data is rare around these target values as depicted in Fig.~\ref{fgr:noraml_multi}. There are some noisy red points in the case of TPSA = 40 and LogP = 2. We suspect that molecules are physically less plausible at the low TPSA and LogP values.

\subsection{Generalization of model on unseen BRICS fragments}

\begin{figure}
 \includegraphics[height=5cm]{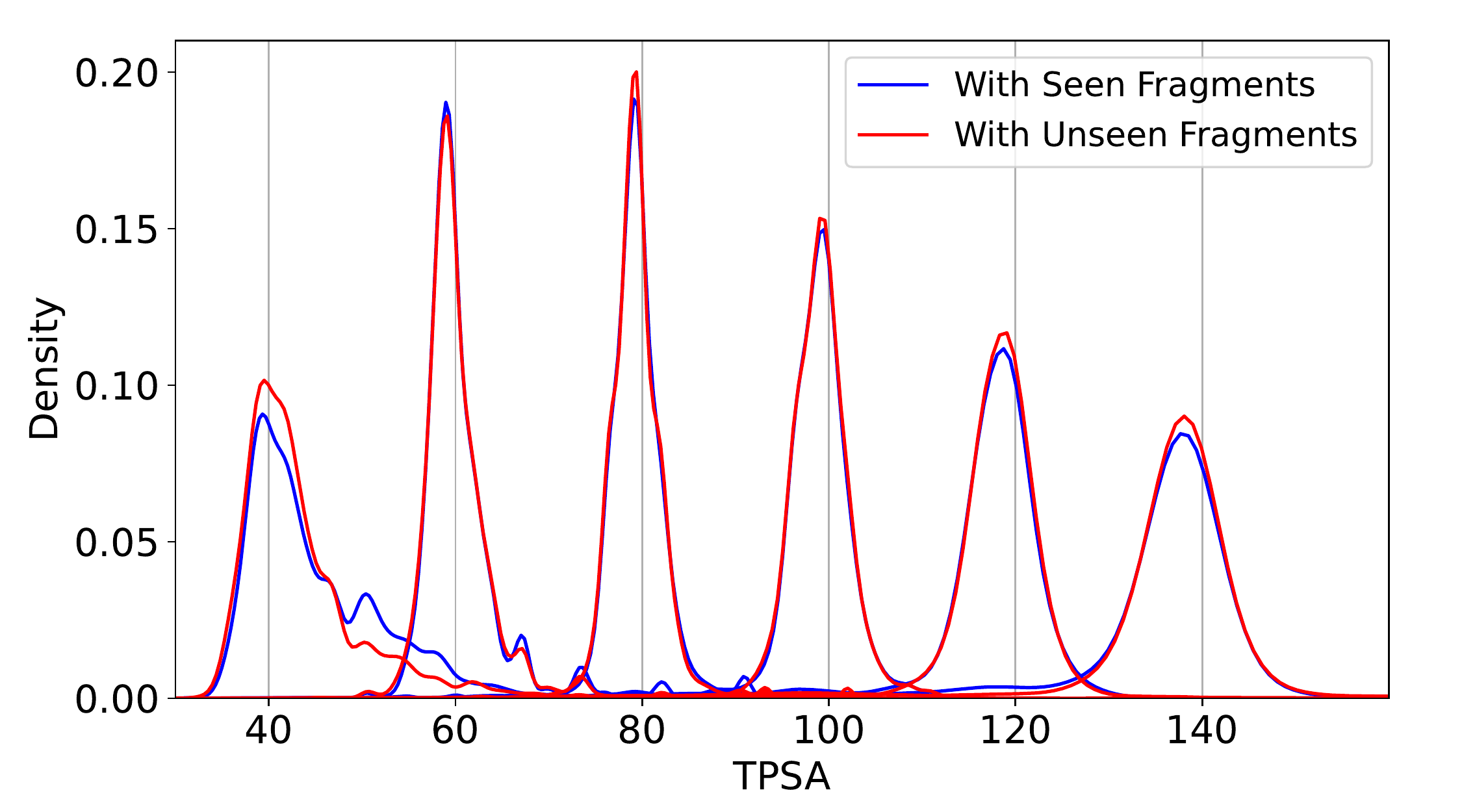}
 \caption{The TPSA distribution of the generated molecules with the seen (blue line) and unseen (red line) BRICS fragments.}
 \label{fgr:compare}
\end{figure}

One distinctive advantage of our model is that any BRICS fragment not necessarily in the training set can be used for molecule generation because the model takes the embedding vectors of BRICS fragments converted by a neural network as input. To verify the feasibility, we tested the model whether it can generate new molecules with target properties at a high success rate by adding unseen BRICS fragments. We first excluded randomly chosen 23,367 BRICS fragments during the training process and used them as the unseen fragments in the generation process. The generation process is identical to that of the previous experiment described in section 3.1 except using only the excluded BRICS fragments for addition. After training the model conditioned on TPSA, we generated 100 molecules using the same 100 starting molecules used in the previous experiments but with the excluded fragments. 
Figure \ref{fgr:compare} shows the result.
The red and blue lines show the property distributions of the generated molecules with the seen and unseen BRICS fragments, respectively. For both cases, the distribution peaks place at the respective target properties, which proves that the model equally well performs with unseen BRICS fragments. 

In the real-world application of molecule generation, it is likely to impose certain constraints in molecule design. For example, in drug discovery, medicinal chemists may want to add a hydrophobic substructure or linker to a core structure according to their structure-activity relationship analysis. The atom addition strategy needs several steps to meet the condition probably with a low success rate. In contrast, our model can readily offer a practical strategy. For instance, we can force the model to add only hydrophilic fragments by providing only hydrophilic BRICS fragments for addition. Our model is particularly suitable for this purpose because it works well with unseen BRICS fragments. 

\begin{figure}
 \includegraphics[height=5cm]{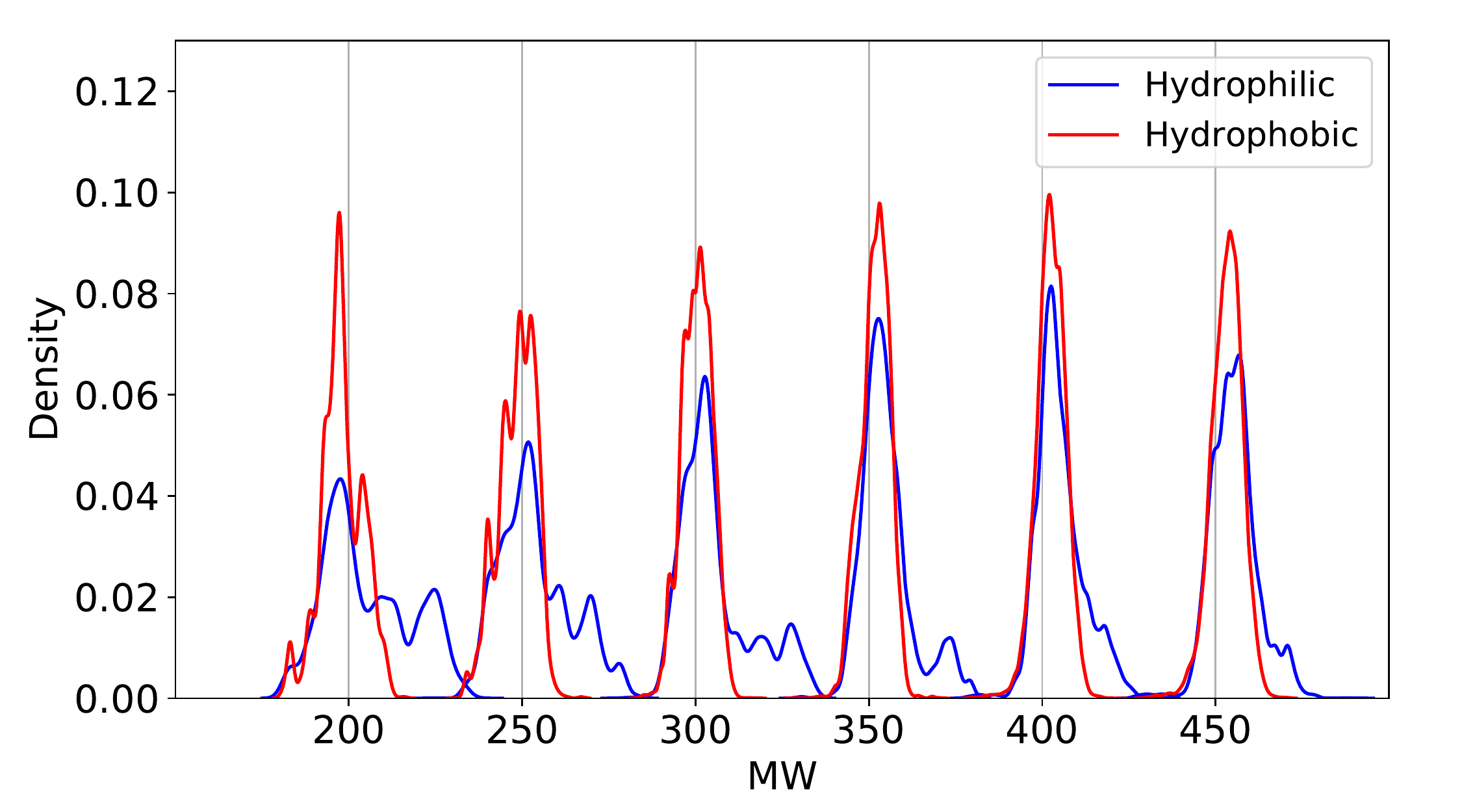}
 \caption{The MW distribution of the generated molecules with the hydrophobic (blue lines) and hydrophilic (red lines) BRICS fragments.}
 \label{fgr:polar_nonpolar_MW.pdf}
\end{figure}

\begin{figure}
 \includegraphics[height=14cm]{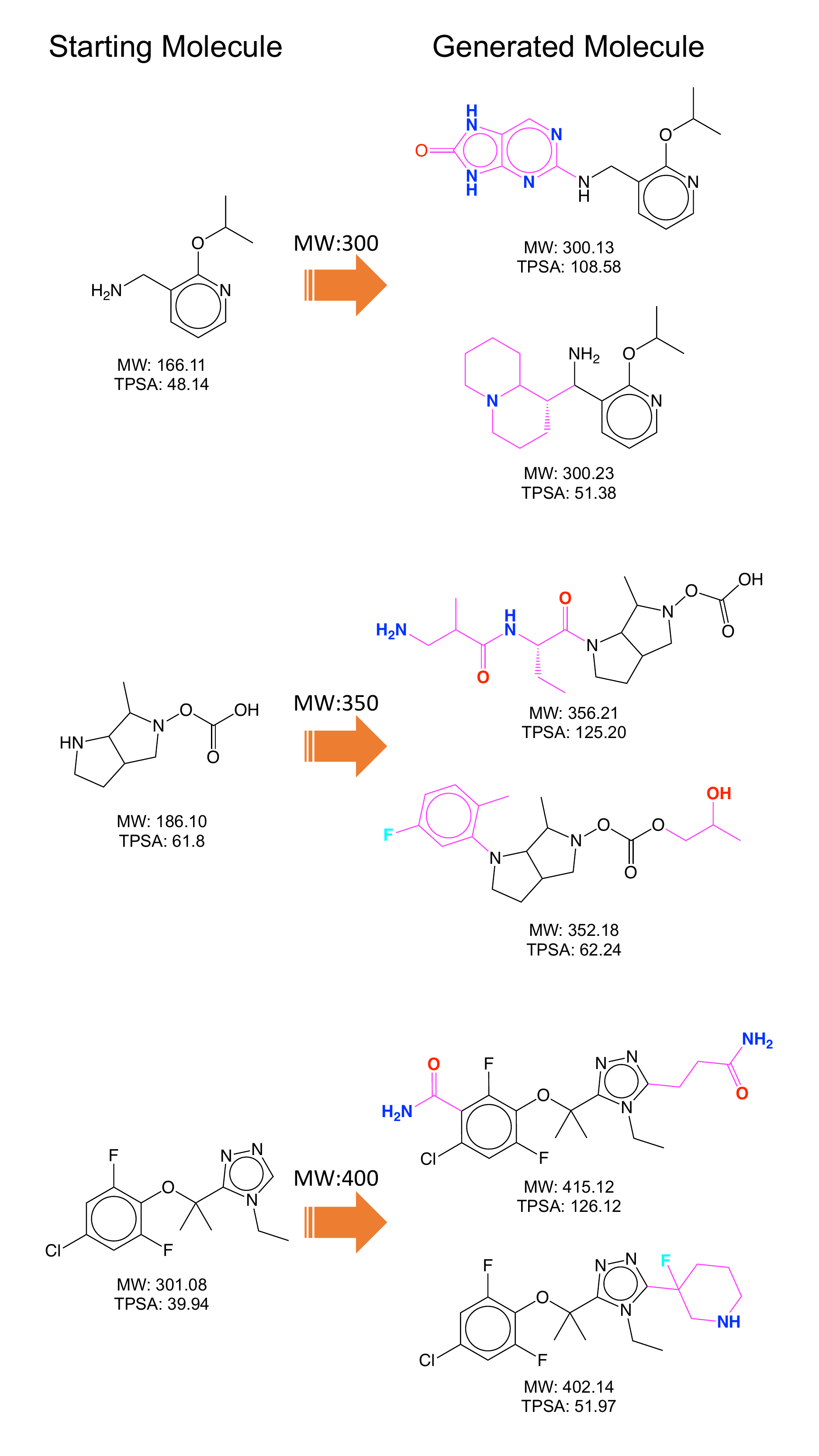}
 \caption{The examples of the generated molecules with the hydrophobic and hydrophilic BRICS fragments. The molecules on the left side are starting molecules. The top and bottom molecules on the right side were obtained from the molecule generation with hydrophilic and hydrophobic BRICS fragments, respectively. The molecules generated with the hydrophilic fragments include more nitrogen and oxygen atoms. In contrast, the molecules generated with the hydrophobic BRICS fragments have more hydrocarbons. }
 \label{fgr:compare_polar_mol}
\end{figure}

For demonstration, we prepared two sets containing unseen hydrophilic and hydrophobic BRICS fragments, respectively. The hydrophobic fragment set is made of 2,000 BRICS fragments with relatively low TPSA values, which appear more than five times in the dataset. For the hydrophilic fragment set, we chose the top 2,000 BRICS fragments in terms of the TPSA value also with more than five times occurrences in the dataset. Fig.~\ref{fgr:polar_nonpolar_MW.pdf} shows the MW distribution of the molecules generated by the model conditioned on MW when we design new molecules with the hydrophilic and hydrophobic fragment sets, respectively. The blue and red lines indicate the generated molecules with hydrophilic and hydrophobic BRICS fragments, respectively. Like the previous result shown in Fig.~\ref{fig:single_property}, each distribution peak locates at the respective target value for both hydrophilic and hydrophobic fragment sets. The distributions of the hydrophilic fragments are slightly more spread than those of the hydrophobic fragments. It is because the hydrophilic set includes relatively larger fragments, making the precise control of MW harder. Fig.~\ref{fgr:compare_polar_mol} shows the several examples of the generated molecules and their starting molecules. The molecules generated with the hydrophilic fragments include more nitrogen and oxygen atoms as intended, whereas those generated with the hydrophobic BRICS fragments have more hydrocarbons. 

\subsection{Application to drug design: demonstration of designing a novel inhibitor against the 3CL protease of SARS-CoV-2}
\begin{figure}
 \includegraphics[width=9cm]{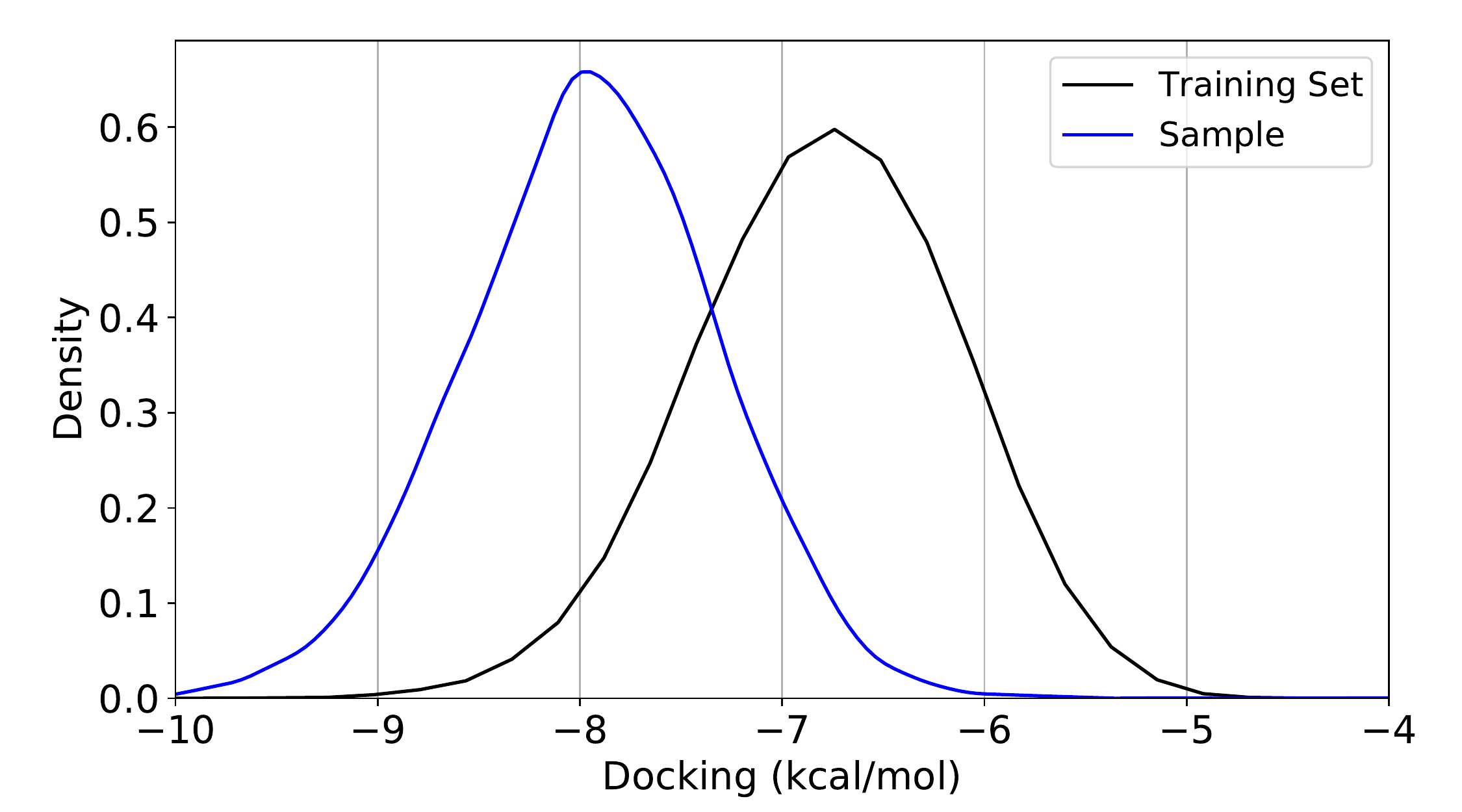}
 \caption{The docking score distributions of the training set and the generated molecules with a target docking score of -9 kcal/mol. The docking score distribution of the generated molecules shifted -1.5 kcal/mol from that of the training set, showing that the generated molecules are more likely to bind to the target protein in terms of the docking score. }
 \label{fgr:docking}
\end{figure}

As a real-world application, we applied our model to designing novel inhibitors against the 3CL protease of SARS-CoV-2. The 3CL protease is one of the most widely studied biological targets for the development of anti-SARS-CoV-2 drug. The objective of this experiment is to let the model learn the relationship between molecular structures and their binding affinities against the target protein. In other words, the model learns what molecular fragments should be added to a given molecule to increase the binding affinity. After training, this model produces molecules that are more likely to bind to the target. Considering more than millions of learnable parameters in the model, we need a large amount of data for training. However, experimental data is not sufficient. Instead, we used simulation data as a demonstration. 
We performed docking calculations with the same molecular library used in the previous experiments in section 3.1 and the 3CL protease of SARS-CoV-2 whose PDB id is 7L13. We used Smina\cite{Koes2013}, the fork of Autodock Vina\cite{Trott2009}, for the docking calculations with the default setting. The initial conformers of the molecules in the library were obtained by the universal force field\cite{Rappe1992} calculation of RDKit. The calculated docking scores have used as the labeled data for conditional generation. Note that the lower the docking score, the higher the binding affinity, due to the negative sign of the docking score. Fig.~\ref{fgr:docking} shows the docking score distribution of the molecules in the training set. The molecules with docking scores lower than -9.0 kcal/mol are in the top 0.64~\% of the training set.  

The generation started with fragments sampled randomly from the test set, yielding 100,000 molecules with a target docking score of -9.0 kcal/mol. Fig.~\ref{fgr:docking} compares the docking score distribution of the generated molecules with that of the training set. The distribution of the generated molecules substantially shifts toward lower docking scores from that of the training set. As a result, the portion of molecules with docking scores lower than -9.0 kcal/mol increased about eight times from 0.64~\% to 4.85~\%, which manifests the feasibility of our model for practical applications to drug design. 

Despite the success of designing molecules with the target docking score to some extents, the distribution peak deviates significantly from the target value, which is contrast to the previous cases of targeting MW, LogP, and TPSA.
MW, LogP, and TPSA can be determined solely by the properties of each fragment for addition. In contrast, the docking score is determined not by the fragment itself but by its interaction with the target protein. Even a single fragment addition may change the entire interaction mode of a molecule by altering its binding pose.
Since we did not consider the target protein structure explicitly in the generation process, the model needs to learn all possible interaction changes upon the fragment addition, which is not straightforward. Therefore, the success rate of designing molecules with a target docking core is substantially lower than that of the previous cases.

\section{Conclusions}
Molecular fragments such as functional groups are closely related to molecular properties and synthetic accessibility. Thus, it is expected that fragment-based molecular design can facilitate better controlling the target properties of generated molecules with a high generalization ability and substantially improving their synthetic accessibility. Here, we proposed a novel fragment-based deep generative model. 
The model generates new molecules by sequentially adding molecular fragments to a given starting molecule. Dealing with many fragments including unseen is essential for a high diversity of the resulting molecules. This cannot be achieved if representing fragments with a fixed vector and applying a classifier to the vector. Instead, we devised a model that predicts the bonding probability of any two molecules: one from the given core molecule and the other from a fragment library. Therefore, the model does not limit the number of fragments in the library. Furthermore, the model takes the embedding vector of fragments encoded by a deep neural network as input, which enables the model to accept unseen fragments after training. This strategy leads to a high generalization ability of the model in term of fragment diversity. Fragments for training and generation were prepared using the BRICS decomposition method which explicitly takes into account synthetic feasibility when decomposing complete molecules. Hence, the model can implicitly learn synthetic accessibility from the data prepared as such.

Our model consists of three modules: fragment selection, atom selection , and termination prediction modules. The fragment selection module evaluates the bonding probability between a given molecule and a molecular fragment. Then, the atom selection module finds out the most probable atom pairs for making a chemical bond between the two molecules. Finally, the termination prediction gives the probability of terminating the molecule generation process. 

We assessed the model performance in various tasks. First, the model was able to control the molecular weight, topological polar surface area, and LogP while generating molecules with a high success rate. Such high performance retained with unseen core molecules or unseen fragments. This result supports that the model achieved a good generalization ability to some extents rather than simply memorizing the hidden pattern of the training set. This generalization ability comes from learning how to tune molecular properties by adding appropriate fragments in a step-wise manner. For instance, the model can design molecules with target properties out of the distribution of the training set. We also demonstrated that target properties can be achieved by adding only hydrophobic or hydrophilic fragments. These cases cannot be done with previous models that generate molecules by adding atoms and bonds. As a practical application, we successfully designed candidate inhibitors showing high binding affinities against the 3CL protease of SARS-CoV-2 in terms of docking score. 
We believe that our fragment-based deep generative model paves a practical way of molecular design with high generalization ability and synthetic accessibility for various chemical applications such as drug discovery.

\section{Reference}

\section*{Author Contributions}
Conceptualization: J.L.; Methodology: J.L., S.S.; Software, Investigation and Formal Analysis: J.L., S.S.; Writing -- Original Draft: J.L.\ and S.S.; Writing -- Review \& Editing: J.L., S.S., and W.Y.K.; Supervision: W.Y.K.

\section*{Conflicts of interest}
There are no conflicts to declare.

\section*{Acknowledgements}
This work was supported by the Tech Incubator Program for Startup (TIPS) funded by the Ministry of SMEs and Startups (MSS, Korea) (S3031674).



\balance


\bibliography{rsc} 
\bibliographystyle{rsc} 

\end{document}